\documentclass[twoside,leqno,twocolumn]{article} 

\usepackage[letterpaper]{geometry}

\usepackage{ltexpprt}
\usepackage[pdfversion=1.5]{hyperref}
\usepackage{graphicx}
\usepackage{amsmath}
\usepackage{amssymb}
\usepackage{booktabs}

\usepackage{tikz}
\usepackage{color}
\usepackage{colortbl}
\usepackage{etex}
\usepackage{morefloats}

\newcommand{\ours}[1]{\textsc{IntervMerge}}

\definecolor{customPurple}{HTML}{9467bd}
\definecolor{customOrange}{HTML}{ff7f0e}
\definecolor{customPink}{HTML}{e377c2}
\definecolor{customBlue}{HTML}{17becf}

\definecolor{lightblue}{HTML}{CCD9FF}
\definecolor{gray}{HTML}{F2F2F2}
\definecolor{green}{HTML}{c6ffb3}
\newcommand{\grayblock}{\raisebox{-.01ex}{\tikz{\node[fill=gray,draw=black,minimum size=0.7em,inner sep=0pt] {};}}\hspace{0.0em}}
\newcommand{\greenblock}{\raisebox{-.01ex}{\tikz{\node[fill=green,draw=black,minimum size=0.7em,inner sep=0pt] {};}}\hspace{0.0em}}
\newcommand{\fullposition}{\greenblock\greenblock\greenblock \greenblock\greenblock\greenblock\greenblock\greenblock\greenblock\greenblock\greenblock\greenblock}
\newcommand{\everytwoeven}{\greenblock\grayblock\greenblock\grayblock\greenblock\grayblock\greenblock\grayblock\greenblock\grayblock\greenblock\grayblock}
\newcommand{\everythree}{\greenblock\grayblock\grayblock\greenblock\grayblock\grayblock\greenblock\grayblock\grayblock\greenblock\grayblock\grayblock}
\newcommand{\everyfour}{\greenblock\grayblock\grayblock\grayblock\greenblock\grayblock\grayblock\grayblock\greenblock\grayblock\grayblock\grayblock}
\newcommand{\threelast}{\grayblock\grayblock\grayblock\grayblock\grayblock\grayblock\grayblock\grayblock\grayblock\greenblock\greenblock\greenblock}
\newcommand{\twolast}{\grayblock\grayblock\grayblock\grayblock\grayblock\grayblock\grayblock\grayblock\grayblock\grayblock\greenblock\greenblock}
\newcommand{\lastlayer}{\grayblock\grayblock\grayblock\grayblock\grayblock\grayblock\grayblock\grayblock\grayblock\grayblock\grayblock\greenblock}
\newcommand{\layereight}{\grayblock\grayblock\grayblock\grayblock\grayblock\grayblock\grayblock\greenblock\grayblock\grayblock\grayblock\grayblock}
\newcommand{\layerfour}{\grayblock\grayblock\grayblock\grayblock\greenblock\grayblock\grayblock\grayblock\grayblock\grayblock\grayblock\grayblock}
\newcommand{\layerzero}{\greenblock\grayblock\grayblock\grayblock\grayblock\grayblock\grayblock\grayblock\grayblock\grayblock\grayblock\grayblock}

\usepackage[capitalize]{cleveref}
\crefname{section}{Sec.}{Secs.}
\Crefname{section}{Section}{Sections}
\Crefname{table}{Table}{Tables}
\crefname{table}{Tab.}{Tabs.}

\begin{document}
\newcommand\relatedversion{}
\renewcommand\relatedversion{\thanks{The full version of the paper can be accessed at \protect\url{https://arxiv.org/abs/1902.09310}}} 

\title{\Large Parameter-Efficient Interventions for Enhanced Model Merging}
\author{Marcin Osial\thanks{Jagiellonian University, IDEAS NCBR}
\and Daniel Marczak\thanks{Warsaw University of Technology, IDEAS NCBR}
\and Bartosz Zieliński\thanks{Jagiellonian University, IDEAS NCBR} }

\date{}

\maketitle


\fancyfoot[R]{\scriptsize{Copyright \textcopyright\ 2025 by SIAM\\
Unauthorized reproduction of this article is prohibited}}





\begin{abstract} \small\baselineskip=9pt
Model merging combines knowledge from task-specific models into a unified multi-task model to avoid joint training on all task data. However, current methods face challenges due to representation bias, which can interfere with tasks’ performance. As a remedy, we propose \ours{}, a novel approach to multi-task model merging that effectively mitigates representation bias across the model using task-specific interventions. To further enhance its efficiency, we introduce mini-interventions, which modify only part of the representation, thereby reducing the additional parameters without compromising performance. Experimental results demonstrate that \ours{} consistently outperforms the state-of-the-art approaches using fewer parameters.
\end{abstract}

\section{Introduction}
\label{sec:intro}

Multi-task learning (MTL) assumes simultaneous accommodation of knowledge from multiple tasks using a common backbone~\cite{caruana1997multitask, vandenhende2021multi}.
This approach offers several benefits, including better performance, eliminating the need for separate task-specific models, and facilitating knowledge transfer between related tasks~\cite{wu2020understanding}.  
However, this approach is challenging, as it can lead to negative transfer or task interference. Furthermore, MTL necessitates collecting data from all tasks for joint training, which can be costly in terms of larger model size, extended training times, and usage of computational resources. Additionally, data privacy concerns may complicate the implementation of MTL, especially in sensitive domains. MTL also lacks the flexibility to incorporate new tasks without retraining the entire model \cite{yu2024unleashing}.

Model merging has emerged as a promising alternative to traditional joint MTL to address these limitations~\cite{li2023deep}. Model merging aims to fuse the abilities of multiple task-specific models into a single multi-task model while keeping the parameter size the same as that of the individual models. Practitioners now follow a standard workflow where they fine-tune a pre-trained foundational model for various target tasks. This approach corresponds to model merging more than traditional multi-task learning. Consequently, large language models (LLMs) have effectively utilized the benefits of this procedure \cite{jin2022dataless, wortsman2022model, xiao2023lm, goddard2024arcee, yu2024language, akiba2024evolutionary, rame2024warp, rame2024rewarded}.
\begin{figure}[t!]
    \centering
    \includegraphics[width=0.47\textwidth]{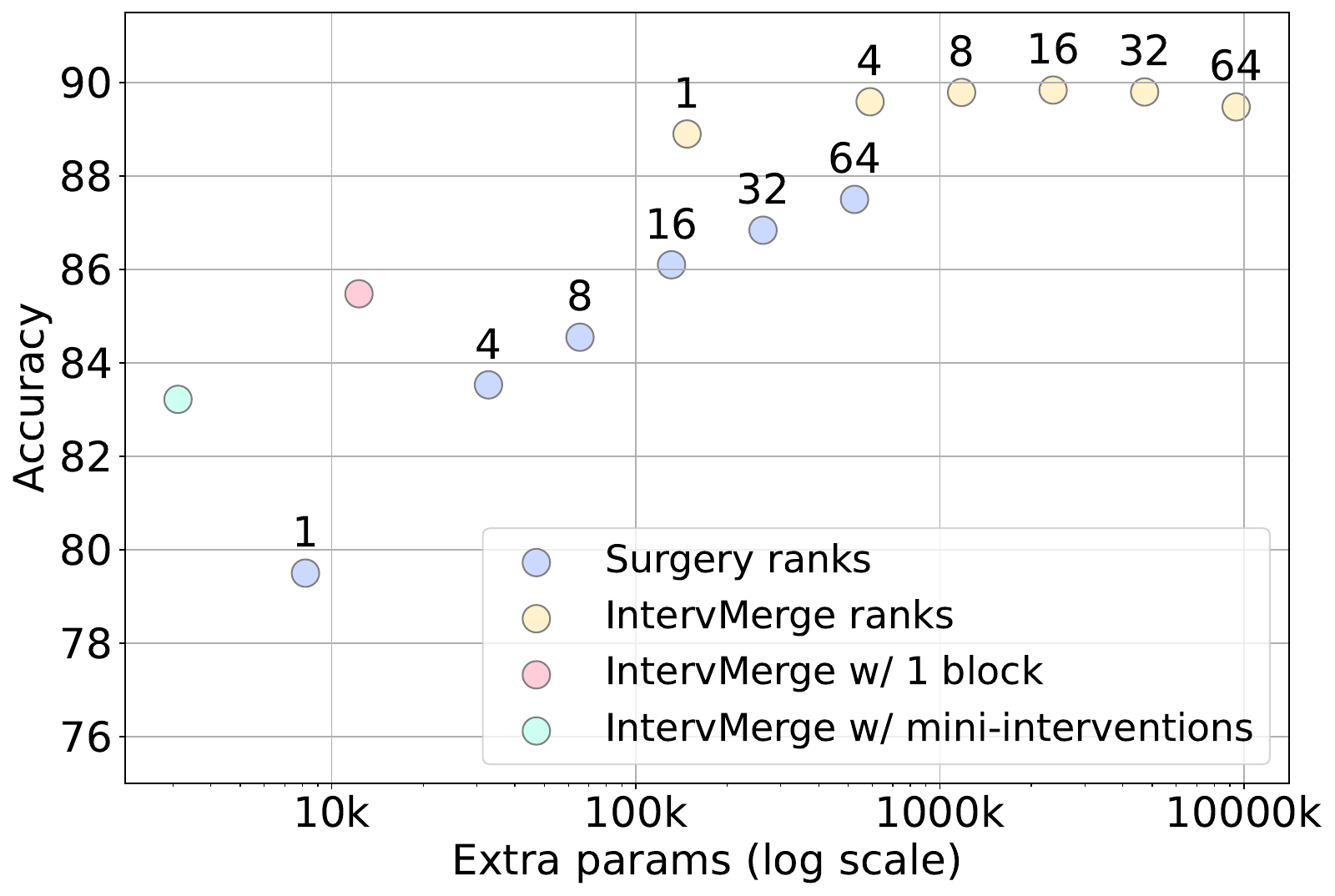}
    \caption{\ours{} consistently demonstrates superior performance compared to the state-of-the-art Surgery approach in multi-task model merging. This advantage is particularly evident when utilizing our efficient mini-intervention mechanism, which achieves better results than Surgery while employing three times fewer parameters. It is important to note that \ours{} may exhibit more additional parameters than Surgery for certain ranks, as it is applied across many network blocks.}   
    \vspace{-0.9em}
    \label{fig:param_plot}
\end{figure}
Merging offers several advantages. It eliminates the need to collect and manage data from all tasks, reducing training costs and data privacy concerns. Model merging enables greater flexibility in incorporating new tasks, supporting continual learning \cite{marczak2024magmax}.
Furthermore, model merging can enhance the robustness and improve the handling of distribution shifts \cite{rame2023model, wortsman2022robust}. To achieve these benefits, merging methods interpolate between the parameters of task-specific models~\cite{ilharco2022patching, ilharco2023task, wortsman2022model}. 
Hybrid methods such as AdaMerging~\cite{yangadamerging} aim to increase merging performance by assuming access to a tiny subset of data from all tasks to optimize the merging parameters.

Several challenges hinder their performance relative to traditional MTL, primarily due to representation bias~\cite{yang2024representation}. This bias occurs when the merged model's feature representations diverge from those of individual task-specific models, leading to task interference and reduced merged model quality. Also, merging methods often struggle to integrate knowledge from different sources effectively, and the complexity of merging models increases with the number of tasks.
Existing solutions, such as Surgery, address the post-merging stage, where the model has initially merged regarding current methods. Surgery attempts to mitigate representation bias by focusing narrowly on the network's output, but it lacks a comprehensive understanding of the interplay of biases across different layers and scenarios. 

In this paper, we introduce \ours{}, an approach designed to comprehensively mitigate representation bias across multiple layers of the network, thereby enhancing the stability and consistency of representations across diverse tasks. Unlike the Surgery method, which addresses bias exclusively at the final layer before classification heads, our approach extends bias mitigation to earlier layers of the network, thereby facilitating a deeper understanding of the nature and nuances of representation biases. 
Furthermore, by concentrating various types of task-specific knowledge at the network's end, Surgery increases representation complexity. This strategy does not guarantee the stability of the merged representations. It also fails to prevent the propagation of errors originating from the initial layers.
In contrast, \ours{} employs efficient mini-intervention operations that facilitate the distribution of task-specific knowledge throughout the network, thereby preventing the escalation of errors. The method \ours{} utilizes task-specific interventions inspired by the ReFT framework~\cite{wu2024reft}. This approach enables adaptive and consistent representation modifications tailored to each task. Each mini-intervention can adjust its rank size, token position, intervention function, and location within the layer depth. Additionally, it can operate on varying lengths and segments of representation. As a result, \ours{} demonstrates superior performance to the Surgery method while utilizing fewer additional parameters, as illustrated in Figure~\ref{fig:param_plot}, by providing an efficient and holistic alignment representation.
\ours{}, is evaluated for image classification tasks across eight diverse datasets, such as SUN397, Stanford Cars, and RESISC45, which can be framed also as a multi-domain learning scenario.

To sum up, our contributions are as follows:
\begin{itemize}
    \item We propose \ours{}, a method of comprehensively reducing representation bias in post-merged models using flexible, task-specific interventions.
    \item We introduce a novel and parameter-efficient mechanism called mini-intervention, which significantly decreases the additional parameters used by the merged model.
    \item We demonstrate that the specific tasks largely influence the misalignment of merged representations, and we identify multiple strategies to achieve improved alignment.
    \item We offer an adaptable solution with multiple options that can be combined to enhance different merging techniques, facilitating effective applications for real-world merging.
\end{itemize}

\section{Related work}


\paragraph{Model merging for Multi-Task Learning.}
The fundamental principle underlying model merging is the concept of linear mode connectivity, which posits that independently trained neural networks with similar architectures often converge to solutions that lie within the same basin of the loss landscape \cite{frankle2020linear, garipov2018loss, draxler2018essentially, wortsman2021learning}. This property allows for meaningful
interpolation between model weights, forming the basis for various merging techniques \cite{foret2020sharpness, dimitriadis2023pareto, sanh2021multitask, neyshabur2020being, wang2024localizing}. The simplest form of model merging is weight averaging as in Model Soup~\cite{wortsman2022model}, where corresponding parameters from different models are directly averaged. When trained on the same task, it improves the accuracy and robustness of the model~\cite{wortsman2022model, lu2022improving, rame2022diverse, izmailov2018averaging}. Multiple single-task models have recently been merged to create a unified multi-task model~\cite{davari2023model}. Notably, Task Arithmetic~\cite{ilharco2023task} utilizes task vectors, representing the differences between the parameters of fine-tuned models and those of pre-trained models. TIES-Merging~\cite{yadav2024ties} addresses conflicts by manipulating these task vectors and resolving parameter redundancies. Further advancements include DARE (Drop And REscale) \cite{yu2024language}, which sparsifies delta parameters, and AdaMerging \cite{yangadamerging}, which learns task-specific merging coeﬀicients per source model or layer of source models with entropy minimalization. Other approaches, e.g., linearizes the fine-tuning process to make the single-task models easier to merge~\cite{ortizjimenez2023tangent}.
A new line of research tries to repair the models created by existing merging techniques to reduce further the performance gap caused by representation bias. 
Surgery~\cite{yang2024representation} trains a set of task-specific adapters that modify the final representations to mitigate the representation bias. 

\section{Method}
\subsection{Preliminaries}
\paragraph{Notation:} Let $f: \mathcal{X} \times \Theta \rightarrow \mathcal{Y}$ represent a neural network function. 
\begin{itemize}
\item $\mathcal{X} \subset \mathbb{R}^d$ represents the $d$-dimensional input space, containing the input samples. 
\item $\Theta \subset \mathbb{R}^m$ denotes the $m$-dimensional parameter space, which encompasses all the learnable parameters of the neural network. 
\item $\mathcal{Y} = \{1, \ldots, c\}$ is the $c$-class output space, where the neural network produces its predictions. 
\end{itemize} The neural network model $f$ is composed of $L$ distinct layers, each with its own set of parameters $\theta^l \in \mathbb{R}^{m_l}$, where $l \in \{1, 2, \ldots, L\}$ and $m = \sum_{l=1}^L m_l$. We can represent the full set of model parameters as $\theta = \{\theta^1, \theta^2, ..., \theta^L\} \in \Theta$. We denote the network as consisting of $N$ blocks, where each block is indexed by $b$, such that $b \in \{1, 2, \ldots, N\}$. For any input $x_i \in \mathcal{X}$, the neural network model $f$ produces an output $y_i = f(x_i, \theta) \in \mathcal{Y}$, which corresponds to the predicted class label and  $i$ denotes the $i$-th sample.

We consider a multi-task learning scenario, with $T$ distinct tasks, each with its task-specific dataset $\mathcal{D}_t$. For each task $t \in \{1, 2, \ldots, T\}$, we have a corresponding task-specific model $f_{\theta_t}$, where $\theta_t$ represents the parameters of that model and ${\theta_{\text{MTL}}}$ denote the final merged multi-task model parameters. Importantly, all the task-specific models $f_{\theta_t}$ are fine-tuned from a standard pre-trained model,
which is initialized with the same set of parameters $\theta_{PRE}$. 
The dataset $\mathcal{D}_t$ for each task is further divided into a training set $\mathcal{D}_t^{tr}$ and a test set $\mathcal{D}_t^{te}$, such that $\mathcal{D}_t = \mathcal{D}_t^{tr} \cup \mathcal{D}_t^{te}$. The task-specific models $f_{\theta_t}$ are trained only on the corresponding training sets $\mathcal{D}_t^{tr}$, without access to the test data $\mathcal{D}_t^{te}$.

\paragraph{Problem Setup:}
The model merging problem aims to combine the weights $\{\theta_t\}_{t=1}^T$ to obtain a final parameters ${\theta_{\text{MTL}}}$, such that ${\theta_{\text{MTL}}}$ can simultaneously perform all $T$ tasks. Crucially, this merging process must be accomplished without access to the original training data $\{\mathcal{D}_t^{tr}\}_{t=1}^T$. The objective is to minimize the loss of the merged model $f_{\theta_{\text{MTL}}}$ on the test datasets $\{\mathcal{D}_t^{te}\}_{t=1}^T$ across all tasks:
\begin{equation}
\min_{\theta_{\text{MTL}}} \frac{1}{T} \sum_{t=1}^T \frac{1}{|\mathcal{D}_t^{te}|} \sum_{(x_i, y_i) \in \mathcal{D}_t^{te}} \ell(f_{\theta_{\text{MTL}}}(x_i), y_i),
\label{eq:min_objective}
\end{equation}
where $\ell(\cdot)$ denotes an appropriate loss function.

\begin{figure*}[t!]
    \centering
    \includegraphics[width=0.99\textwidth]{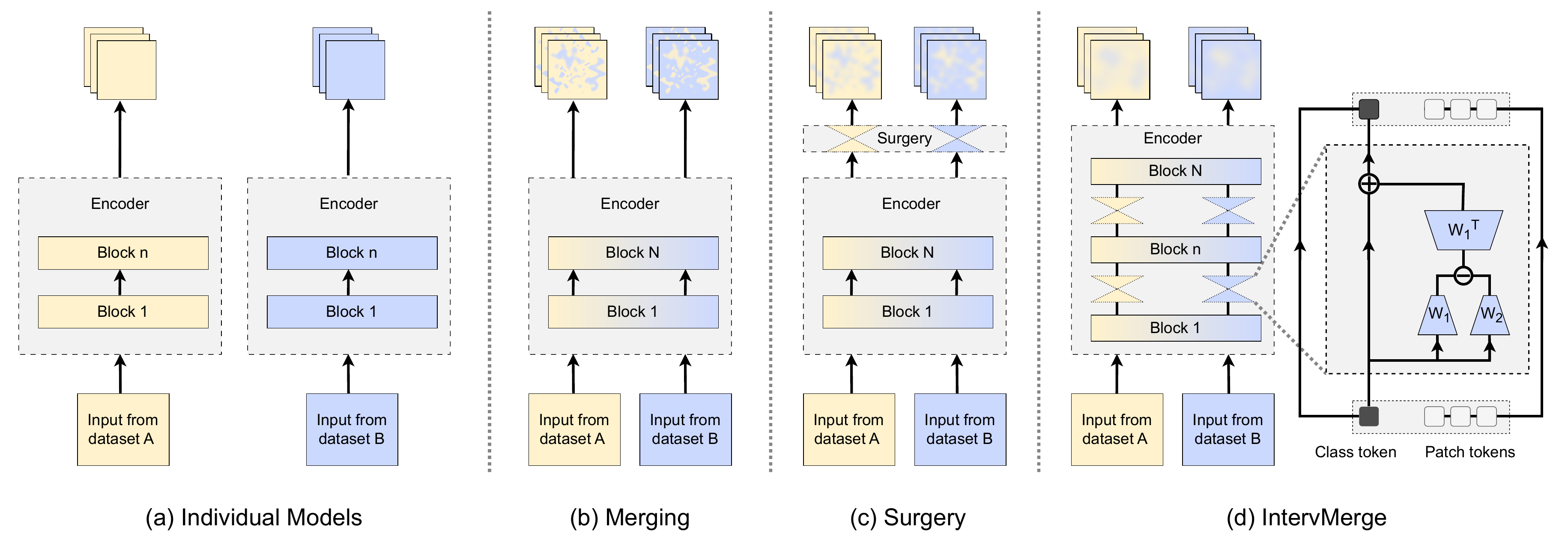}
    \caption{Various solutions of MTL have different issues. Multiple individually trained models (a) require storing and serving separate weights for each task. Traditional model merging (b) schemes combine multiple individual models into one but often lead to performance degradation. Surgery (c) addresses the problem of representational bias but only on the final layer of the encoder. Our \ours{} (d) aims to overcome those limitations by applying lightweight interventions across the whole network, mitigating interference between tasks.}
    \label{fig:method}
    \vspace{-0.5em}
\end{figure*}

\paragraph{Baseline Merging Solutions:} Recent approaches to effectively model merging have focused on various strategies to combine task-specific models. Weight Averaging~\cite{utans1996weight, wortsman2022model} simply calculates the mean of model parameters across tasks: $\theta_{\text{MTL}} = \frac{1}{T} \sum_{t=1}^T \theta_t$. Task Arithmetic~\cite{ilharco2023task} introduces the concept of task vectors, defined as the difference between fine-tuned and pre-trained weights $\tau_t = \theta_t - \theta_{\text{PRE}}$, and merges these vectors with a scaling factor: $\theta_{\text{MTL}} = \theta_{\text{PRE}} + \lambda \sum_{t=1}^T \tau_t$. Ties-Merging~\cite{yadav2023tiesmerging} builds upon this by applying additional operations to resolve conflicts between task vectors: $\theta_{\text{MTL}} = \theta_{\text{PRE}} + \lambda \sum_{t=1}^T \phi(\tau_t)$, where $\phi(\cdot)$ represents trimming, sign election and disjoint merge. AdaMerging~\cite{yangadamerging} further refines this approach by learning adaptive coefficients for merging, either at the task level: $\theta_{\text{MTL}} = \theta_{\text{PRE}} + \sum_{t=1}^T \lambda_t \tau_t$, or the layer level:
$\theta_{\text{MTL}} = \theta_{\text{PRE}} + \sum_{l=1}^L \sum_{t=1}^T \lambda_{t}^l \tau_{t}^l$, where $\lambda_{t}^l$ represents the adaptive coefficient for task $t$ at layer $l$, and $\tau_t^l$ denotes the task vector for task $t$ at layer $l$.

Building upon these direct merging methods, Surgery~\cite{yang2024representation} takes a complementary approach by improving an already merged model. To address the representation bias, task-specific modules that modify the final representations of the merged model are introduced. For each task $t$, it applies a transformation: $h_t^{S} = \mathbf{W}_t^{up} \text{ReLU}(\mathbf{W}_t^{down} h_t)$, where $\mathbf{W}_t^{down}$ and $\mathbf{W}_t^{up}$ are task-specific down- and up-projection matrices, respectively. Here, $h_t^{S}$ is the output after the Surgery module, while $h_t$ denotes the representation obtained after the encoder. Let \( g_{\theta_t}(x_i) \) represent the whole Surgery merged model with task-specific adapters. Let  $\ell(\cdot)$ denote the L1 loss function, then distillation loss is given by:
\begin{equation}
\mathcal{L}_{\text{distill}} = \frac{1}{T} \sum_{t=1}^T \ell(f_{\theta_t}(x_i), g_{\theta_t}(x_i))
\label{eq:distill}
\end{equation}


\subsection{\ours{}}
Our method initiates with a model \( f_{\theta_{\text{MTL}}} \) that has already been merged using established techniques, such as AdaMerging, akin to the Surgery approach. To enhance the stability and consistency of the merged representations, we introduce task-specific lightweight modules $\Phi_b^t: \mathbb{R}^k \rightarrow \mathbb{R}^k$ at every block $b$ for each task $t$ within the network. These modules actively refine the representations throughout the architecture, as illustrated in Figure~\ref{fig:method}(d). Our approach employs the same distillation loss as in Equation \ref{eq:distill}, but we utilize intervention modules in exchange for Surgery adapters.

We integrate the task-specific modules $\Phi_b^t$ into the standard architecture of Vision Transformer (ViT) blocks. We position each module after the Multi-Head Self-Attention (MHSA) operation and before the subsequent residual connection. The complete sequence of operations within a single transformer block $b$ for task $t$, incorporating \ours{}, can be formalized as:
\begin{equation}
\begin{aligned}
z_b &= \text{MHSA}(\text{LN}(h_b)) \\
h'_b &= h_b + \Phi_b^t(z_b) \\
h_{b+1} &= h'_b + \text{MLP}(\text{LN}(h'_b)),
\end{aligned}
\label{eq:operation_seq}
\end{equation}
where $h_b$ is the input representation to the $b$-th transformer block. The task-specific modules $\Phi_b^t$ primarily operate on the representation of the \texttt{[CLS]} token only.

\subsubsection{Full-Intervention} 

Our lightweight, task-specific modules draw inspiration from the ReFT approach~\cite{wu2024reft}, initially designed for language models and trained with standard cross-entropy loss. For our specific use case in model merging, we adapt the ReFT functions accordingly. The following form of intervention is selected for block $b$ and task $t$:
\begin{equation}
\Phi_b^t(z_b) = z_b + \mathbf{W}_2^T (\mathbf{W}_1 z_b + \mathbf{b} - \mathbf{W}_2 z_b),
\label{eq:full_intervention}
\end{equation}
where $\mathbf{W}_1, \mathbf{W}_2 \in \mathbb{R}^{k \times r}$ are low-rank, non-orthogonal, learnable projection matrices, and $\mathbf{b}$ is a bias vector associated with $\mathbf{W}_1$. The representation dimension is denoted by $k$, while $r$ denotes the rank of the projection.

Figure~\ref{fig:method}(d) illustrates how these modules are integrated into the overall architecture. Early adjustments in the network can mitigate representation errors before they propagate, reducing the need for more extensive modifications in later layers. The effectiveness of these improvements may vary depending on the rank $r$ chosen, as different biases and their locations within the network can influence the extent of the required adjustments.

\begin{figure}[h!]
    \centering
    \includegraphics[width=0.25\textwidth]{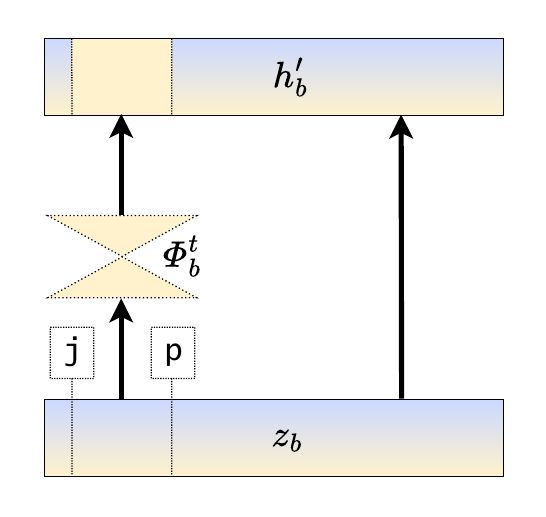}
    \caption{Illustration of the mini-intervention approach, where specific part $[j\!:\!p]$ of the representation $z_b$ is modified by intervention $\Phi_b^t$ to produce the updated representation $h'_b$.}
    \vspace{-1.0em}
    \label{fig:mini_intervention}
\end{figure}

\subsubsection{Mini-Intervention}
We propose a novel mini-intervention approach for parameter-efficient intervention. This method modifies only selected representation positions at each layer, as presented in Figure ~\ref{fig:mini_intervention}. The core idea of our mini-intervention mechanism for task $t$ and block $b$ can be formalized as:
\begin{equation}
\begin{split}
h_b' &= h_b + \Phi_b^t(z_b{[j:p]}),
\end{split}
\label{eq:part_intervention}
\end{equation}
where $z_b$ is the full representation, $z_b{[j:k]}$ represents a part of representation from index $j$ to $p$. The indices $j$ and $p$ may be fixed or determined dynamically based on the layer index, total representation dimension, or other heuristics. This focused modification strategy facilitates precise control over the representation, effectively incorporating new information while maintaining the integrity and contextual coherence of the original embedding.

\section{Experimental Setup}

\paragraph{Datasets:} Following the experimental setup of recent multi-tasks model merging approaches studies~\cite{yang2024representation, yangadamerging, ilharco2023task}, we evaluate our method on eight diverse image classification tasks: SUN397~\cite{xiao2016sun}, Stanford Cars~\cite{krause20133d}, RESISC45~\cite{cheng2017remote}, EuroSAT~\cite{helber2019eurosat}, SVHN~\cite{netzer2011reading}, GTSRB~\cite{stallkamp2011german}, MNIST~\cite{lecun1998mnist}, and DTD~\cite{cimpoi2014describing}. The datasets encompass various categories and contexts, including natural scenes, fine-grained vehicle models, and satellite imagery for remote sensing, extending to digit recognition tasks. 
\paragraph{Baselines:} We compare our method against several approaches:
\begin{itemize}
    \item Non-merging: Pre-trained, Individual, and Traditional MTL.
    \item Standard model merging: Weight Averaging~\cite{wortsman2022model}, Task Arithmetic~\cite{ilharco2023task}, Ties-Merging~\cite{yadav2023tiesmerging}, and AdaMerging~\cite{yangadamerging}. 
    \item Post-merging:
    Our primary point of comparison is the state-of-the-art Surgery~\cite{yang2024representation}.
\end{itemize}

\paragraph{Architectures:} In line with recent work in model merging~\cite{yang2024representation, yangadamerging, yadav2023tiesmerging}, we evaluate our method on two variants of the CLIP~\cite{radford2021learning} visual encoder: ViT-B/32 and ViT-L/14, with findings for the latter included in the Supplement due to space limitations.

\paragraph{Hyperparameters.}
Our method includes several essential hyperparameters. By default, the total number of transformer blocks $b$ where interventions are applied is $N$ and set to 12. The rank of the low-rank projections, $r$, is set to 1 for most experiments unless specified otherwise. All other hyperparameters, including the number of iterations, batch size, optimizer, and lambda values, were set to match those used in the Surgery method (see Supplement). 

\section{Results}
By default, we present averaged results for eight datasets for all Tables and use the intervention function denoted in Equation \ref{eq:full_intervention}. When reporting results with standard deviation ($\pm$), these are calculated based on three independent runs with different seeds. When integrating our method with previous merging approaches that use lambda $\lambda$ parameters, we jointly optimize these lambdas along with our intervention modules, allowing for greater flexibility. For a fair comparison, we also provide results where we train the Surgery with learnable lambdas (see Table~\ref{tab:diff_ranks}).

\begin{figure}[t!]
    \centering    
    \includegraphics[width=0.46\textwidth]{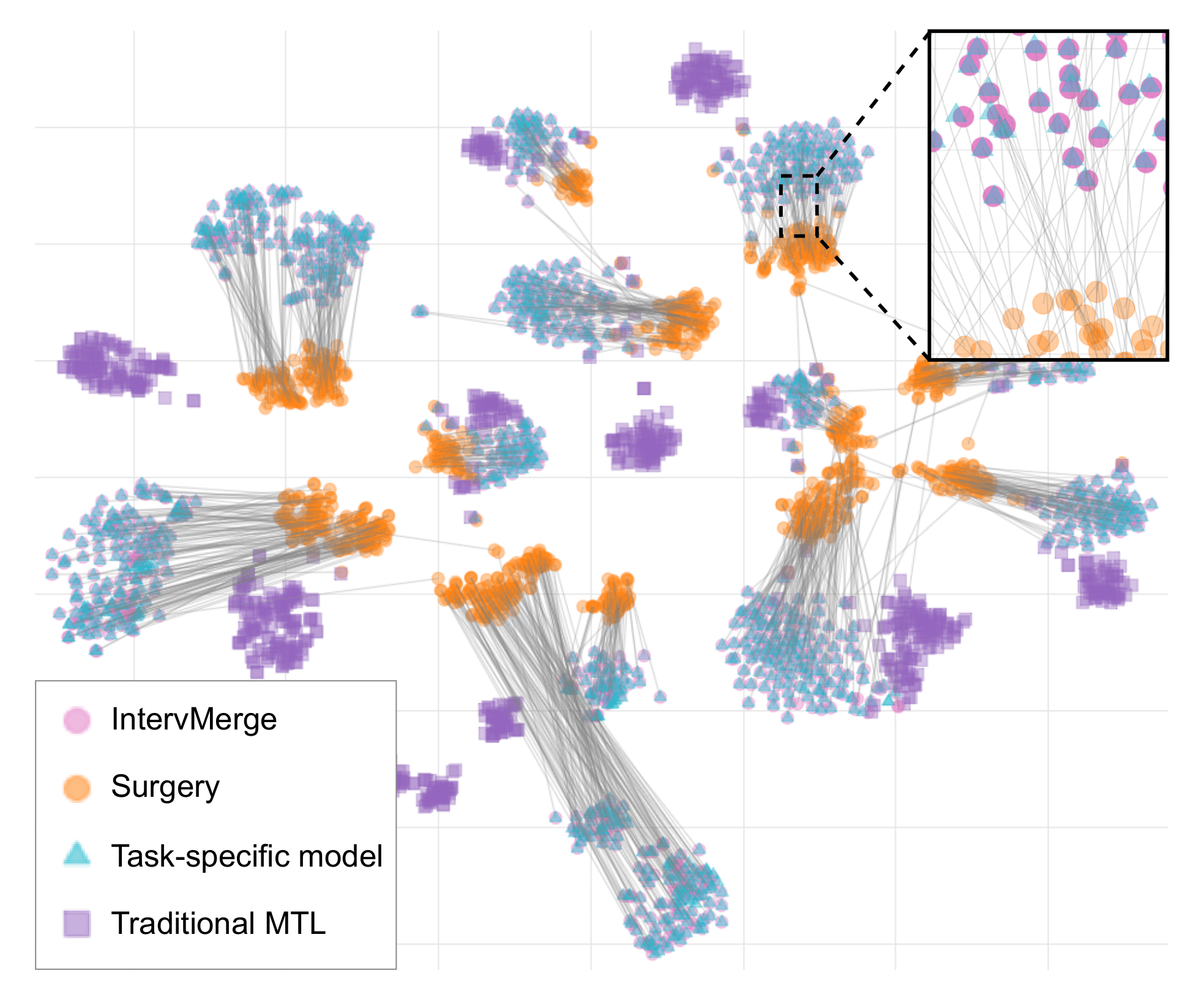}
    \caption{Representations of various methods obtained for the RESISC45 dataset. The gray lines connect individual samples between Surgery and \ours{}. Consequently, the representations of \ours{} are closer to those obtained by the task-specific model.}
    \label{fig:tsne_after_encoder_resisc}
    \vspace{-0.5em}
\end{figure}

\begin{figure}[t!]
    \centering
    \includegraphics[width=0.46\textwidth]{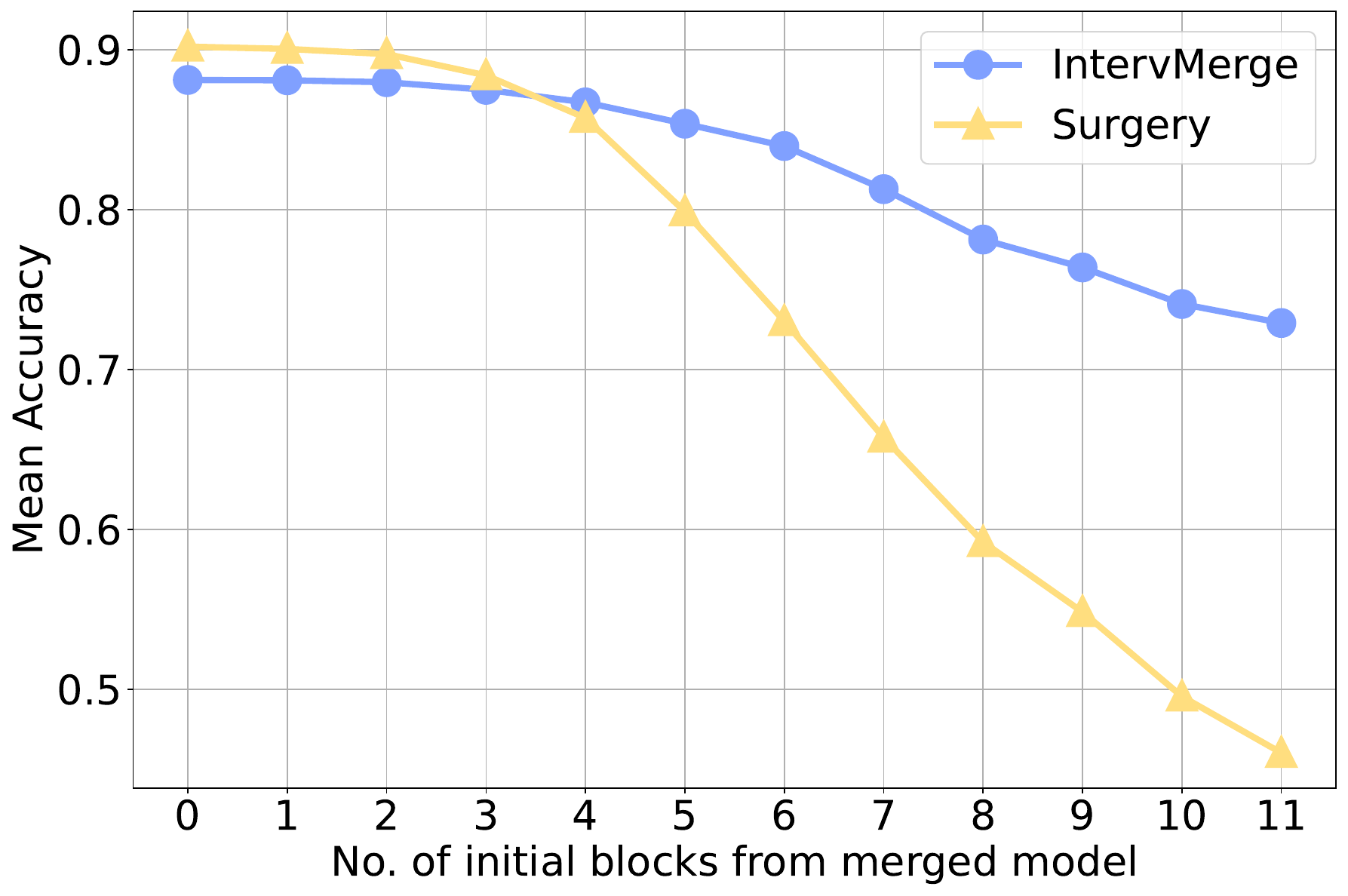}
    \caption{Utilizing a stitched network demonstrates that \ours{} achieves significantly higher accuracy than the Surgery method, reflecting improved consistency with task-specific representations. We averaged the results across eight datasets.}
    \vspace{-0.5em}
    \label{fig:move_to_finenued}
\end{figure}

\paragraph{Comprehensive Mitigation of Representation Bias}
The final encoder outputs of \ours{} exhibit significantly reduced representation bias. Figure~\ref{fig:tsne_after_encoder_resisc} qualitatively illustrates this improvement, showing that the representations generated by \ours{} cluster more closely with those of the task-specific models than the Surgery model. Furthermore, representations of \ours{} are more consistent and stable than Surgery, shown using the cross-task linearity~\cite{zhou2024cross}. We construct a network that integrates the initial blocks from the merged model with the remaining layers from the task-specific model. As illustrated in Figure~\ref{fig:move_to_finenued}, \ours{} maintains its accuracy significantly better than Surgery.
\begin{table*}[h!]
\centering
\scalebox{0.66}{
\begin{tabular}{c|c|cccccccc|c}
\toprule
\textbf{Method} & \textbf{Extra params} & \textbf{SUN397} & \textbf{Cars} & \textbf{RESISC45} & \textbf{EuroSAT} & \textbf{SVHN} & \textbf{GTSRB} & \textbf{MNIST} & \textbf{DTD} & \textbf{Avg.} \\ 
\midrule
Pre-trained & - & 62.3 & 59.7 & 60.7 & 45.5 & 31.4 & 32.6 & 48.5 & 43.8 & 48.0  \\
Individual & - & 75.3 & 77.7 & 96.1 & 99.7 & 97.5 & 98.7 & 99.7 & 79.4 & 90.5 \\ 
Traditional MTL & - & 73.9 & 74.4 & 93.9 & 98.2 & 95.8 & 98.9 & 99.5 & 77.9 & 88.9 \\ \midrule
Weight Averaging & 0 & 65.3 & 63.4 & 71.4 & 71.7 & 64.2 & 52.8 & 87.5 & 50.1 & 65.8 \\
Weight Averaging w/ Surgery & 131k & 67.6 & 64.6 & 85.8 & 96.8 & 76.9 & 82.9 & 97.8 & 67.3 & 80.0 \\
\rowcolor{lightblue}
Weight Averaging w/ \ours{} &  147k & 68.19 & 61.99 & 88.11 & 98.81 & 93.56 & 85.83 & 98.75 & 65.27 & \textbf{82.56} \\ \midrule
Task Arithmetic & 0 & 55.2 & 54.9 & 66.7 & 78.9 & 80.2 & 69.7 & 97.3 & 50.4 & 69.1 \\
Task Arithmetic w/ Surgery & 131k & 63.8 & 59.9 & 83.3 & 97.9 & 87.0 & 87.0 & 98.6 & 69.4 & 80.9 \\
\rowcolor{lightblue}
Task Arithmetic w/ \ours{} &  147k & 69.32 & 65.14 & 90.56 & 99.19 & 95.77 & 94.89 & 99.37 & 69.52 & \textbf{85.47} \\ \midrule
Ties-Merging & 0 & 65.0 & 64.4 & 74.8 & 77.4 & 81.2 & 69.3 & 96.5 & 54.5 & 72.9 \\
Ties-Merging w/ Surgery & 131k & 69.8 & 66.1 & 87.3 & 97.5 & 86.7 & 87.6 & 98.5 & 71.6 & 83.1 \\
\rowcolor{lightblue}
Ties-Merging w/ \ours{} & 147k & 70.08 & 69.02 & 87.54 & 97.78 & 95.07 & 92.95 & 99.23 & 62.71 & \textbf{84.3} \\ \midrule
Taskwise AdaMerging & 0 & 58.0 & 53.2 & 68.8 & 85.7 & 81.1 & 84.4 & 92.4 & 44.8 & 71.1 \\
TW AdaMerging w/ Surgery & 131k & 63.9 & 57.6 & 84.2 & 98.2 & 87.6 & 92.7 & 98.0 & 66.8 & 81.1 \\
\rowcolor{lightblue}
TW AdaMerging w/ \ours{} & 147k & 66.0 & 68.93 & 92.13 & 98.74 & 96.63 & 96.16 & 99.22 & 71.12 & \textbf{86.12} \\ \midrule
AdaMerging & 0 & 64.5 & 68.1 & 79.2 & 93.8 & 87.0 & 91.9 & 97.5 & 59.1 & 80.1 \\
AdaMerging w/ Surgery & 131k & 69.8 & 71.0 & 88.9 & 98.1 & 91.7 & 96.5 & 98.8 & 73.6 & 86.1 \\
AdaMerging w/ Surgery ($r=64$) & 524k & 71.2 & 72.0 & 92.3 & 99.0 & 92.2 & 97.9 & 99.0 & 76.1 & 87.5 \\
\rowcolor{lightblue}
AdaMerging w/ \ours{} & 147k & 70.24 & 75.3 & 94.11 & 99.52 & 96.32 & 98.49 & 99.52 & 78.19 & \textbf{88.96} \\
AdaMerging w/ \ours{} ($r=4$) & 590k & 72.14 & 75.84 & 94.86 & 99.56 & 96.78 & 98.71 & 99.48 & 79.57 & 89.62 \\ 
AdaMerging w/ \ours{} (1 block) & 12k & 66.08 & 72.8 & 87.37 & 99.07 & 94.56 & 96.29 & 99.11 & 68.51 & 85.48 \\
AdaMerging w/ \ours{} (mini-interv.) & 3k & 65.17 & 73.11 & 82.52 & 98.37 & 93.17 & 92.57 & 98.94 & 63.72 & 83.45 \\
\bottomrule
\end{tabular}
}
\caption{Performance comparison of various merging methods and our \ours{} approach using ViT-B/32 models. It can be observed that incorporating \ours{} into various merging methods outperforms incorporating Surgery, even with fewer parameters. Please note that the results display accuracy (\%) for a single run of \ours{} with rank 1 and Surgery with rank 16 unless otherwise specified. Moreover, the extra parameters are counted solely for inference.}
\label{tab:multitask_performance_vitb32}
\end{table*}

\paragraph{Multi-task Model Merging}
Table~\ref{tab:multitask_performance_vitb32} presents the results showcasing the superior performance of \ours{} method compared to Surgery across various model merging techniques. When we apply \ours{} with rank 1 to AdaMerging, it surpasses the previous state-of-the-art Surgery with rank 64 by $1.46\%$, using $3.5$ times fewer parameters. Our method demonstrates substantial gains across other merging techniques, achieving improvements of 2.56\% over Weight Averaging, 4.57\% over Task Arithmetic, and a significant 5.02\% gain over Task-wise AdaMerging.

By applying interventions with rank 1 to a single block, we reduce the parameter count compared to Surgery while maintaining performance improvements of 85.48\%. We developed a vary-length mini-intervention approach that applies interventions across all network blocks. To demonstrate the significant efficiency of the 64-part mini-intervention, we implemented an experiment using the '$h + \mathbf{R}^T(\mathbf{b})$' formula with rank 1. It achieves an average accuracy 3\% higher than Surgery, using fewer parameters (as illustrated in Figure~\ref{fig:param_plot}). The parameter increase from \ours{} mini-intervention is minimal compared to the overall model size.

\subsection{Selecting Token for Intervention}
Our experiments with rank $r=4$ interventions, reveal that the positioning of interventions within the token sequence affects model performance to varying degrees (Table~\ref{tab:position_performance}). Notably, applying the intervention to the class token achieves the highest average accuracy (89.49\%), outperforming all other configurations.
As this study focuses on image classification, our findings reinforce the established importance of the class token in Vision Transformers for aggregating global information~\cite{dosovitskiy2020vit}. However, in other type of visual tasks, the relevance of patch tokens may differ, suggesting the need for a tailored approach to token selection.

\begin{table}[t!]
\centering
\scalebox{0.9}{
\begin{tabular}{c|c}
\midrule
\textbf{Token with Intervention} & \textbf{Accuracy} \\ \midrule
first token & $88.79 \pm 0.08$ \\
middle token & $88.57 \pm 0.10$\\
last token & $88.79 \pm 0.04$\\
\rowcolor{lightblue}
class token & $\mathbf{89.49 \pm 0.04}$ \\
patch tokens & $70.82 \pm 0.33$ \\
all tokens & $88.92 \pm 0.06$ \\
\bottomrule
\end{tabular}
}
\caption{The effectiveness of the intervention varies depending on the token to which it is applied. The highest accuracy is achieved when intervening on class tokens.}
\label{tab:position_performance}
\vspace{-1.0em}
\end{table}

\subsection{Blocks with Intervention}

\begin{table}[t!]
\centering
\scalebox{0.85}{
\begin{tabular}{c|c|c}
\midrule
\textbf{No. of blocks} & \textbf{Accuracy} & \textbf{Blocks with intervention} \\ 
\midrule
\rowcolor{lightblue}
12 & $\mathbf{88.83}$ & \fullposition \\ 
6 & 88.13 & \everytwoeven \\ 
4 & 87.80 & \everythree \\ 
3 & 87.20 & \everyfour \\ 
3 & 85.65 & \threelast \\ 
2 & 84.44 & \twolast \\ 
1 & 79.50 & \lastlayer \\ 
1 & 83.82 & \layereight \\ 
1 & 85.48 & \layerfour \\ 
1 & 84.67 & \layerzero \\ 
\bottomrule
\end{tabular}
}
\caption{Optimal performance is attained when interventions are applied across all blocks. Nevertheless, similarly strong results can be achieved by intervening every second block. Notice that the blocks with intervention are marked in green.}
\label{tab:layerwise_analysis}
\vspace{-0.5em}
\end{table}

Table~\ref{tab:layerwise_analysis} presents the results of our experiments varying on the number and position of blocks to which we apply the interventions. We observe that applying interventions to all 12 blocks yields the highest average accuracy (88.83\%), and a decrease in performance is observed when we reduce the number of blocks. 

Interestingly, when we restrict interventions to only a subset of blocks, we see that the choice of blocks matters. For instance, applying interventions to every fourth block (3 blocks total) achieves an accuracy of 87.20\%, notably higher than applying interventions to only the last three blocks (85.65\%). This investigation suggests that distributed interventions across the network are more effective than concentrated interventions near the output. 

When limiting interventions to a single block, we observe that the choice of block significantly impacts performance. Editing the middle blocks (85.48\%) tends to yield better results than editing only the first (84.67\%) or last (79.50\%) block. This experiment suggests that middle blocks are crucial in refining task-specific representations on average for these eight tasks. It's important to note that the optimal block for single-block interventions is task-specific. For instance, our analysis (see the Supplement) revealed that for tasks like SUN397, the effectiveness of interventions decreased from one of the last blocks towards the first. Conversely, for the GTSRB, editing the last block decreased performance from 92\% to 85\% compared to the base merged model.

\subsection{Mini-Intervention}

Table~\ref{tab:miniedits} presents the results examining the impact of the intervention for different parts and of the representation across all 12 blocks with default formula (Equation \ref{eq:full_intervention}). While the average accuracy does not vary dramatically across different configurations, some notable trends exist.

Interventions applied to the latter parts of the representation yield slightly better performance. For instance, when editing 256 elements, the last part (512-768) achieves the highest accuracy of 87.77\%. This pattern is consistent across different intervention lengths.

\begin{figure}[t!]
    \centering    
    \includegraphics[width=0.45\textwidth]{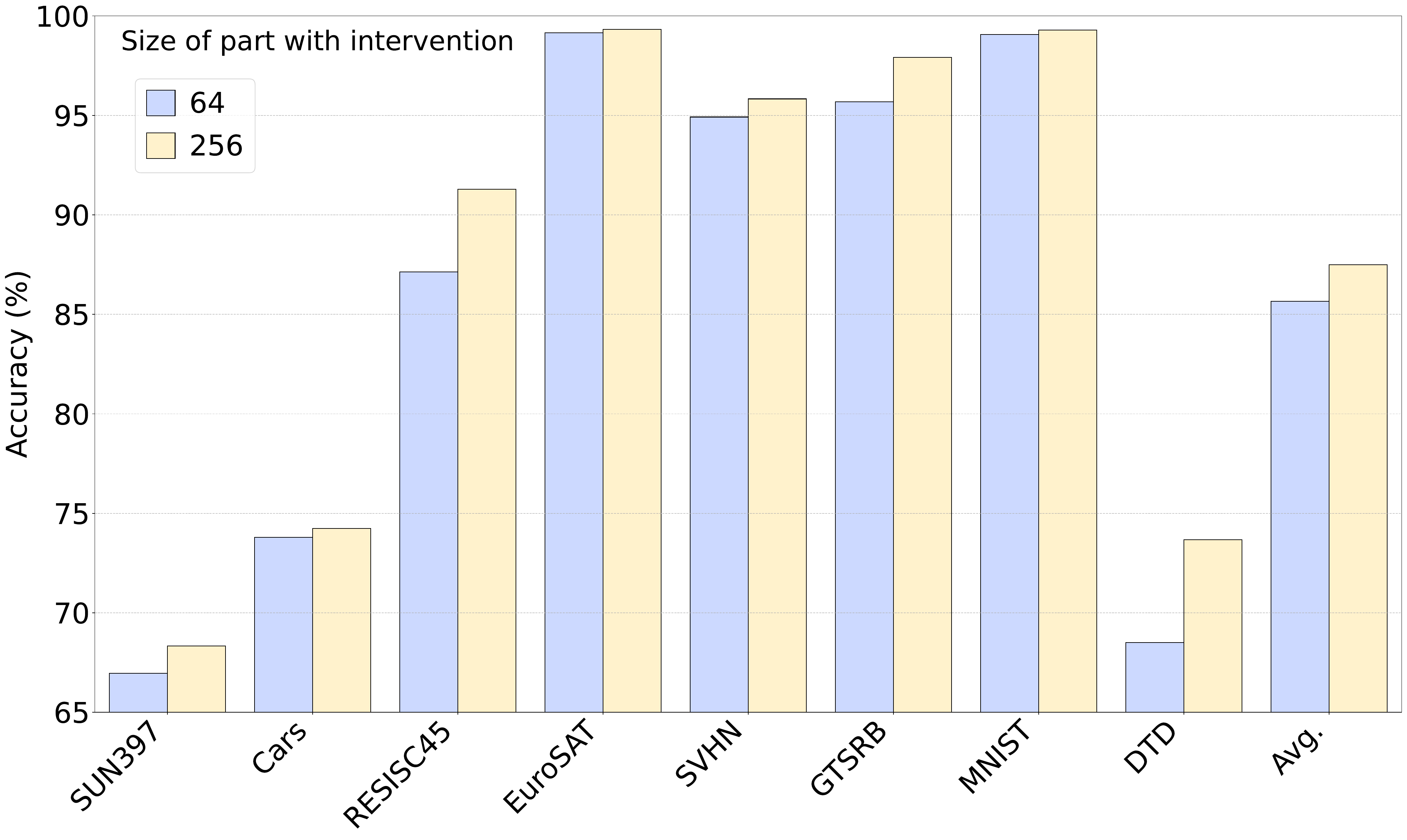}
    \caption{Accuracy across considered datasets fluctuates with the size of the intervened part, generally yielding higher results for larger parts. However, outcomes differ significantly depending on the task.}
    \vspace{-1.0em}
    \label{fig:miniedit_len_diff}
\end{figure}

The effectiveness of applying interventions to different part sizes varies across tasks, generally tending to decrease as the edit length is reduced. However, the situation is more nuanced and task-dependent. For instance (see Figure~\ref{fig:miniedit_len_diff}), EuroSAT shows only a 0.3\% change between 256 and 64-element edits, while DTD exhibits a significant 6\% difference.

Interestingly, applying edits to unique positions in each successive block (shifting by 64 elements per block) produces the best outcomes, achieving 85.63\% accuracy for 64-element edits. These results suggest that the position of edits within the representation can have subtle but meaningful effects on model performance.

\begin{table}[t!]
\centering
\scalebox{0.9}{
\begin{tabular}{@{}c|c|c@{}}
\toprule
\textbf{Part with Intervention} & \textbf{Part size} & \textbf{Accuracy} \\ 
\midrule
0-256 & 256 & $87.59 \pm 0.06$ \\
256-512 & 256 & $87.54 \pm 0.07$\\
\rowcolor{lightblue}
512-768 & 256 & $\mathbf{87.77 \pm 0.03}$\\ \midrule
0-128 & 128 & $86.69 \pm 0.03$ \\ 
320-448 & 128 & $86.66 \pm 0.03$ \\
\rowcolor{lightblue}
640-768 & 128 & $\mathbf{86.73 \pm 0.04}$\\ \midrule
0-64 & 64 & $85.44 \pm 0.05$ \\
352-416 & 64 & $85.22 \pm 0.04$ \\ 
704-768 & 64 & $85.35 \pm 0.03$\\ 
0-64, 64-128, (shift) & 64 & $\mathbf{85.63 \pm 0.03}$ \\
\midrule
0-32 & 32 & $83.95 \pm 0.01$\\
\rowcolor{lightblue}
368-400 & 32 & $\mathbf{84.10 \pm 0.03}$\\
736-768 & 32 & $84.06 \pm 0.03$\\
\bottomrule
\end{tabular}
}
\caption{Accuracy depends primarily on the size of the representation part to which we apply the mini-intervention. The longer, the better. Moreover, it is preferred to shift the part position across blocks.}
\label{tab:miniedits}
\end{table}

\begin{table}[t!]
\centering
\scalebox{0.8}{
\begin{tabular}{@{}c|c|c@{}}
\toprule
\textbf{Part with intervention} & \textbf{No. of blocks} & \textbf{Accuracy} \\ 
\midrule
0-128 & 6 & $85.36 \pm 0.02$ \\ \midrule
\rowcolor{lightblue}
0-64 (with shift) & 12 &$\mathbf{85.63 \pm 0.03}$ \\
0-32, 736-768 & 12 & $85.36 \pm 0.01$\\
0-16, 200-184, 568-584, 752-768  & 12 & $85.21 \pm 0.02$ \\ \midrule
0-786 (best) & 1 & $85.42 \pm 0.06$ \\ 
0-786 (worst)& 1 & $79.41 \pm 0.28$ \\
\bottomrule
\end{tabular}
}
\caption{When operating on the same total parameter budget, applying mini-interventions to smaller representation parts across more blocks is preferred over searching for optimal larger parts on fewer blocks.}
\label{tab:param_equal}
\vspace{-1.0em}
\end{table}

Table~\ref{tab:param_equal} compares various edit configurations for \ours{} method using the same total parameter budget. The performance differences across configurations are minimal, with the shifted distributed approach achieving slightly higher accuracy (85.63\%) than optimal single-block edit (85.48\%). These results suggest that task-specific knowledge can be distributed throughout the network and exclusively injected in specific blocks or positions. However, the effectiveness of distributed mini-edits across all blocks indicates that it is more convenient and stable to apply small interventions throughout the network rather than searching for an optimal single block for intervention.
\subsection{Rank of the Intervention}
In Table~\ref{tab:diff_ranks}, we observe that even with weaker merging methods like Weight Averaging and Ties-Merging, \ours{} demonstrates significant improvements as rank increases. For instance, Weight Averaging with \ours{} improves from 82.56\% at rank 1 to 88.3\% at rank 64, surpassing Surgery's performance on more robust merging methods. This highlights the effectiveness of our distributed editing approach in enhancing even basic merging techniques. Furthermore, the performance of \ours{} plateaus or slightly decreases at rank 64 for most methods, suggesting an optimal range for rank selection.
Most significantly, when applied to AdaMerging, \ours{} consistently outperforms traditional MTL (88.9\%) across all ranks. It's worth noting that Surgery with learnable lambdas shows only marginal improvements over the standard Surgery method, indicating the limitations of this approach.
\subsection{Intervention Functions and Data Availability}
Table~\ref{tab:module_operations} presents various intervention functions. The results indicate that these different intervention methods yield comparable performance levels. Some formulations utilize fewer parameters by employing a single orthogonal matrix $\mathbf{R}$. 

Our method demonstrates a clear advantage over Surgery when the data visibility ratio is lower, as illustrated in the table \ref{tab:impact_of_data_availability}.
\begin{table}[t!]
\centering
\scalebox{0.73}{
\begin{tabular}{c|cccc}
\midrule
\textbf{Method} & \multicolumn{4}{c}{\textbf{Rank}} \\
 & 1 & 8 & 16 & 64 \\ \midrule
Ties-Merging w/ \ours{} & 84.3 & 86.58 & 86.86 & 86.91 \\
Weight Averaging w/ \ours{} & 82.56 & 87.13 & 87.64 & 88.3 \\
Task Arithmetic w/ \ours{} & 85.47 & 88.28 & 88.49 & 88.88 \\
TW AdaMerging w/ \ours{} & 86.12 & 88.72 & 89.03 & 88.97 \\
\rowcolor{lightblue}
AdaMerging w/ \ours{}  & 88.90 & 89.79 & \textbf{89.84} & 89.48 \\ \midrule
AdaMerging w/ Surgery & 81.34 & 84.55 & 86.10 & 87.50 \\
AdaMerging w/ Surgery \textsuperscript{\textdagger} & 82.17 & 85.4 & 86.36 & 88.03 \\
\bottomrule
\end{tabular}
}
\caption{Accuracy generally improves with increasing rank. However, even with lower ranks, our method outperforms Surgery. Note that \textdagger{} corresponds to Surgery with learnable lambda coefficients.}
\label{tab:diff_ranks}
\vspace{-0.4em}
\end{table}

\begin{table}[t!]
\centering
\scalebox{0.9}{
\begin{tabular}{c|c}
\midrule
\textbf{Intervention pattern} & \textbf{Accuracy} \\ 
\midrule

$h + \mathbf{R}^T(\mathbf{b})$ & $88.39 \pm 0.04$ \\ 
$h + \mathbf{R}^T(\mathbf{b} - \mathbf{R}h)$ & $88.96 \pm 0.01$ \\ 
$h + \mathbf{R}^T(\mathbf{W}h + \mathbf{b})$ & $89.53 \pm 0.05$ \\ 
\rowcolor{lightblue}
$h + \mathbf{W}_2^T(\mathbf{W}_1h + \mathbf{b} - \mathbf{W}_2h)$ & $\textbf{89.56 $\pm$ 0.02}$ \\ 
$h + \mathbf{R}^T(\mathbf{W}h + \mathbf{b} - \mathbf{R}h)$ & $89.49 \pm 0.03$ \\
\bottomrule
\end{tabular}
}
\caption{Performance comparison of various intervention patterns in the \ours{} framework.}
\label{tab:module_operations}
\vspace{-0.4em}
\end{table}

\begin{table}[h!]
\centering
\scalebox{0.84}{
\begin{tabular}{lcc}
\midrule
\textbf{Method} & \textbf{Available Test Set} & \textbf{Avg.} \\ 
\midrule
AdaMerging w/ Surgery & 1\% & 82.8 \\ 
\rowcolor{lightblue}
AdaMerging w/ \ours{} & 1\% & 85.55 \\ 
\midrule
AdaMerging w/ Surgery & 5\% & 83.8 \\
\rowcolor{lightblue}
AdaMerging w/ \ours{} & 5\% & 87.78 \\ 
\midrule
AdaMerging w/ Surgery & 10\% & 84.7 \\ 
\rowcolor{lightblue}
AdaMerging w/ \ours{} & 10\% & 88.47\\ 
\bottomrule
\end{tabular}
}
\caption{Impact of the amount of available test data with \ours{} with rank 1 full-interventions.}
\label{tab:impact_of_data_availability}
\vspace{-0.4em}
\end{table}
\section{Limitations and conclusions}
While our findings may extend to domains like LLMs, this study is limited to image classification.
We place interventions after the MHSA layer due to its critical role in capturing global dependencies. We have yet to explore their impact when positioned after other layer types. \ours{} offers promising opportunities for adaptation to object detection and segmentation, as interventions can operate in any kind and number of tokens-like representations, and mini-interventions offer multiple options for optimization. We see adaptability for tasks that involve analyzing complete sequences, like short video analysis. Early interference errors caused by merging methods in understanding the initial frames can significantly disrupt the comprehension of the overall narrative. Additionally, \ours{} can be adapted for CNN by unrolling feature maps, allowing for mini-interventions at multiple spatial positions. 

In this work, we introduced \ours{}, a novel approach to addressing representation bias in multi-task model merging.
Our experiments across diverse visual recognition tasks demonstrated that \ours{} consistently outperforms state-of-the-art merging techniques,  offering a comprehensive toolbox and pipeline that extends through the post-merging phase, while using significantly fewer parameters.


{\small
 \bibliographystyle{plain} 
 \bibliography{references}

\begin{thebibliography}{10}

\bibitem{akiba2024evolutionary}
Takuya Akiba, Makoto Shing, Yujin Tang, Qi~Sun, and David Ha.
\newblock Evolutionary optimization of model merging recipes.
\newblock {\em arXiv preprint arXiv:2403.13187}, 2024.

\bibitem{caruana1997multitask}
Rich Caruana.
\newblock Multitask learning.
\newblock {\em Machine learning}, 28:41--75, 1997.

\bibitem{cheng2017remote}
Gong Cheng, Junwei Han, and Xiaoqiang Lu.
\newblock Remote sensing image scene classification: Benchmark and state of the art.
\newblock {\em Proceedings of the IEEE}, 105(10):1865--1883, 2017.

\bibitem{cimpoi2014describing}
Mircea Cimpoi, Subhransu Maji, Iasonas Kokkinos, Sammy Mohamed, and Andrea Vedaldi.
\newblock Describing textures in the wild.
\newblock In {\em Proceedings of the IEEE conference on computer vision and pattern recognition}, pages 3606--3613, 2014.

\bibitem{davari2023model}
MohammadReza Davari and Eugene Belilovsky.
\newblock Model breadcrumbs: Scaling multi-task model merging with sparse masks.
\newblock {\em arXiv preprint arXiv:2312.06795}, 2023.

\bibitem{dimitriadis2023pareto}
Nikolaos Dimitriadis, Pascal Frossard, and Fran{\c{c}}ois Fleuret.
\newblock Pareto manifold learning: Tackling multiple tasks via ensembles of single-task models.
\newblock In {\em International Conference on Machine Learning}, pages 8015--8052. PMLR, 2023.

\bibitem{dosovitskiy2020vit}
Alexey Dosovitskiy, Lucas Beyer, Alexander Kolesnikov, Dirk Weissenborn, Xiaohua Zhai, Thomas Unterthiner, Mostafa Dehghani, Matthias Minderer, Georg Heigold, Sylvain Gelly, Jakob Uszkoreit, and Neil Houlsby.
\newblock An image is worth 16x16 words: Transformers for image recognition at scale.
\newblock {\em ICLR}, 2021.

\bibitem{draxler2018essentially}
Felix Draxler, Kambis Veschgini, Manfred Salmhofer, and Fred Hamprecht.
\newblock Essentially no barriers in neural network energy landscape.
\newblock In {\em International conference on machine learning}, pages 1309--1318. PMLR, 2018.

\bibitem{foret2020sharpness}
Pierre Foret, Ariel Kleiner, Hossein Mobahi, and Behnam Neyshabur.
\newblock Sharpness-aware minimization for efficiently improving generalization.
\newblock {\em arXiv preprint arXiv:2010.01412}, 2020.

\bibitem{frankle2020linear}
Jonathan Frankle, Gintare~Karolina Dziugaite, Daniel Roy, and Michael Carbin.
\newblock Linear mode connectivity and the lottery ticket hypothesis.
\newblock In {\em International Conference on Machine Learning}, pages 3259--3269. PMLR, 2020.

\bibitem{garipov2018loss}
Timur Garipov, Pavel Izmailov, Dmitrii Podoprikhin, Dmitry~P Vetrov, and Andrew~G Wilson.
\newblock Loss surfaces, mode connectivity, and fast ensembling of dnns.
\newblock {\em Advances in neural information processing systems}, 31, 2018.

\bibitem{goddard2024arcee}
Charles Goddard, Shamane Siriwardhana, Malikeh Ehghaghi, Luke Meyers, Vlad Karpukhin, Brian Benedict, Mark McQuade, and Jacob Solawetz.
\newblock Arcee's mergekit: A toolkit for merging large language models.
\newblock {\em arXiv preprint arXiv:2403.13257}, 2024.

\bibitem{helber2019eurosat}
Patrick Helber, Benjamin Bischke, Andreas Dengel, and Damian Borth.
\newblock Eurosat: A novel dataset and deep learning benchmark for land use and land cover classification.
\newblock {\em IEEE Journal of Selected Topics in Applied Earth Observations and Remote Sensing}, 12(7):2217--2226, 2019.

\bibitem{ilharco2023task}
Gabriel Ilharco, Marco~T{\'{u}}lio Ribeiro, Mitchell Wortsman, Ludwig Schmidt, Hannaneh Hajishirzi, and Ali Farhadi.
\newblock Editing models with task arithmetic.
\newblock In {\em ICLR}, 2023.

\bibitem{ilharco2022patching}
Gabriel Ilharco, Mitchell Wortsman, Samir~Yitzhak Gadre, Shuran Song, Hannaneh Hajishirzi, Simon Kornblith, Ali Farhadi, and Ludwig Schmidt.
\newblock Patching open-vocabulary models by interpolating weights.
\newblock In {\em NeurIPS}, 2022.

\bibitem{izmailov2018averaging}
Pavel Izmailov, Dmitrii Podoprikhin, T.~Garipov, D.~Vetrov, and A.~Wilson.
\newblock Averaging weights leads to wider optima and better generalization.
\newblock {\em Conference on Uncertainty in Artificial Intelligence (UAI)}, 2018.

\bibitem{jin2022dataless}
Xisen Jin, Xiang Ren, Daniel Preotiuc-Pietro, and Pengxiang Cheng.
\newblock Dataless knowledge fusion by merging weights of language models.
\newblock {\em arXiv preprint arXiv:2212.09849}, 2022.

\bibitem{krause20133d}
Jonathan Krause, Michael Stark, Jia Deng, and Li~Fei-Fei.
\newblock 3d object representations for fine-grained categorization.
\newblock In {\em Proceedings of the IEEE international conference on computer vision workshops}, pages 554--561, 2013.

\bibitem{lecun1998mnist}
Yann LeCun.
\newblock The mnist database of handwritten digits.
\newblock {\em http://yann. lecun. com/exdb/mnist/}, 1998.

\bibitem{li2023deep}
Weishi Li, Yong Peng, Miao Zhang, Liang Ding, Han Hu, and Li~Shen.
\newblock Deep model fusion: A survey.
\newblock {\em arXiv preprint arXiv: 2309.15698}, 2023.

\bibitem{lu2022improving}
Peng Lu, I.~Kobyzev, Mehdi Rezagholizadeh, Ahmad Rashid, A.~Ghodsi, and P.~Langlais.
\newblock Improving generalization of pre-trained language models via stochastic weight averaging.
\newblock {\em Conference on Empirical Methods in Natural Language Processing (EMNLP)}, 2022.

\bibitem{marczak2024magmax}
Daniel Marczak, Bartłomiej Twardowski, Tomasz Trzciński, and Sebastian Cygert.
\newblock Magmax: Leveraging model merging for seamless continual learning.
\newblock 2024.

\bibitem{netzer2011reading}
Yuval Netzer, Tao Wang, Adam Coates, Alessandro Bissacco, Baolin Wu, and Andrew~Y Ng.
\newblock Reading digits in natural images with unsupervised feature learning.
\newblock In {\em NIPS Workshop on Deep Learning and Unsupervised Feature Learning}, page~4, Granada, Spain, 2011.

\bibitem{neyshabur2020being}
Behnam Neyshabur, Hanie Sedghi, and Chiyuan Zhang.
\newblock What is being transferred in transfer learning?
\newblock {\em Advances in neural information processing systems}, 33:512--523, 2020.

\bibitem{ortizjimenez2023tangent}
Guillermo Ortiz{-}Jim{\'{e}}nez, Alessandro Favero, and Pascal Frossard.
\newblock Task arithmetic in the tangent space: Improved editing of pre-trained models.
\newblock In {\em NeurIPS}, 2023.

\bibitem{radford2021learning}
Alec Radford, Jong~Wook Kim, Chris Hallacy, Aditya Ramesh, Gabriel Goh, Sandhini Agarwal, Girish Sastry, Amanda Askell, Pamela Mishkin, Jack Clark, et~al.
\newblock Learning transferable visual models from natural language supervision.
\newblock In {\em International conference on machine learning}, pages 8748--8763. PMLR, 2021.

\bibitem{rame2023model}
Alexandre Ram{\'e}, Kartik Ahuja, Jianyu Zhang, Matthieu Cord, L{\'e}on Bottou, and David Lopez-Paz.
\newblock Model ratatouille: Recycling diverse models for out-of-distribution generalization.
\newblock In {\em International Conference on Machine Learning}, pages 28656--28679. PMLR, 2023.

\bibitem{rame2024rewarded}
Alexandre Rame, Guillaume Couairon, Corentin Dancette, Jean-Baptiste Gaya, Mustafa Shukor, Laure Soulier, and Matthieu Cord.
\newblock Rewarded soups: towards pareto-optimal alignment by interpolating weights fine-tuned on diverse rewards.
\newblock {\em Advances in Neural Information Processing Systems}, 36, 2024.

\bibitem{rame2024warp}
Alexandre Ram{\'e}, Johan Ferret, Nino Vieillard, Robert Dadashi, L{\'e}onard Hussenot, Pierre-Louis Cedoz, Pier~Giuseppe Sessa, Sertan Girgin, Arthur Douillard, and Olivier Bachem.
\newblock Warp: On the benefits of weight averaged rewarded policies.
\newblock {\em arXiv preprint arXiv:2406.16768}, 2024.

\bibitem{rame2022diverse}
Alexandre Rame, Matthieu Kirchmeyer, Thibaud Rahier, Alain Rakotomamonjy, Patrick Gallinari, and Matthieu Cord.
\newblock Diverse weight averaging for out-of-distribution generalization.
\newblock {\em NeurIPS}, 2022.

\bibitem{sanh2021multitask}
Victor Sanh, Albert Webson, Colin Raffel, Stephen~H Bach, Lintang Sutawika, Zaid Alyafeai, Antoine Chaffin, Arnaud Stiegler, Teven~Le Scao, Arun Raja, et~al.
\newblock Multitask prompted training enables zero-shot task generalization.
\newblock {\em arXiv preprint arXiv:2110.08207}, 2021.

\bibitem{stallkamp2011german}
Johannes Stallkamp, Marc Schlipsing, Jan Salmen, and Christian Igel.
\newblock The german traffic sign recognition benchmark: a multi-class classification competition.
\newblock In {\em The 2011 international joint conference on neural networks}, pages 1453--1460. IEEE, 2011.

\bibitem{utans1996weight}
Joachim Utans.
\newblock Weight averaging for neural networks and local resampling schemes.
\newblock In {\em Proc. AAAI-96 Workshop on Integrating Multiple Learned Models. AAAI Press}, pages 133--138. Citeseer, 1996.

\bibitem{vandenhende2021multi}
Simon Vandenhende, Stamatios Georgoulis, Wouter Van~Gansbeke, Marc Proesmans, Dengxin Dai, and Luc Van~Gool.
\newblock Multi-task learning for dense prediction tasks: A survey.
\newblock {\em IEEE TPAMI}, 2021.

\bibitem{wang2024localizing}
Ke~Wang, Nikolaos Dimitriadis, Guillermo Ortiz-Jimenez, François Fleuret, and Pascal Frossard.
\newblock Localizing task information for improved model merging and compression.
\newblock {\em ICML}, 2024.

\bibitem{wortsman2021learning}
Mitchell Wortsman, Maxwell~C Horton, Carlos Guestrin, Ali Farhadi, and Mohammad Rastegari.
\newblock Learning neural network subspaces.
\newblock In {\em International Conference on Machine Learning}, pages 11217--11227. PMLR, 2021.

\bibitem{wortsman2022model}
Mitchell Wortsman, Gabriel Ilharco, Samir~Ya Gadre, Rebecca Roelofs, Raphael Gontijo-Lopes, Ari~S Morcos, Hongseok Namkoong, Ali Farhadi, Yair Carmon, Simon Kornblith, et~al.
\newblock Model soups: averaging weights of multiple fine-tuned models improves accuracy without increasing inference time.
\newblock In {\em International conference on machine learning}, pages 23965--23998. PMLR, 2022.

\bibitem{wortsman2022robust}
Mitchell Wortsman, Gabriel Ilharco, Jong~Wook Kim, Mike Li, Simon Kornblith, Rebecca Roelofs, Raphael~Gontijo Lopes, Hannaneh Hajishirzi, Ali Farhadi, Hongseok Namkoong, et~al.
\newblock Robust fine-tuning of zero-shot models.
\newblock In {\em Proceedings of the IEEE/CVF conference on computer vision and pattern recognition}, pages 7959--7971, 2022.

\bibitem{wu2020understanding}
Sen Wu, Hongyang Zhang, and Christopher Ré.
\newblock Understanding and improving information transfer in multi-task learning.
\newblock {\em ICLR}, 2020.

\bibitem{wu2024reft}
Zhengxuan Wu, Aryaman Arora, Zheng Wang, Atticus Geiger, Dan Jurafsky, Christopher~D. Manning, and Christopher Potts.
\newblock Reft: Representation finetuning for language models.
\newblock {\em arXiv preprint arXiv: 2404.03592}, 2024.

\bibitem{xiao2016sun}
Jianxiong Xiao, Krista~A Ehinger, James Hays, Antonio Torralba, and Aude Oliva.
\newblock Sun database: Exploring a large collection of scene categories.
\newblock {\em International Journal of Computer Vision}, 119:3--22, 2016.

\bibitem{xiao2023lm}
Shitao Xiao, Zheng Liu, Peitian Zhang, and Xingrun Xing.
\newblock Lm-cocktail: Resilient tuning of language models via model merging.
\newblock {\em arXiv preprint arXiv:2311.13534}, 2023.

\bibitem{yadav2023tiesmerging}
Prateek Yadav, Derek Tam, Leshem Choshen, Colin Raffel, and Mohit Bansal.
\newblock {TIES}-merging: Resolving interference when merging models.
\newblock In {\em NeurIPS}, 2023.

\bibitem{yadav2024ties}
Prateek Yadav, Derek Tam, Leshem Choshen, Colin~A Raffel, and Mohit Bansal.
\newblock Ties-merging: Resolving interference when merging models.
\newblock {\em Advances in Neural Information Processing Systems}, 36, 2024.

\bibitem{yang2024representation}
Enneng Yang, Li~Shen, Zhenyi Wang, Guibing Guo, Xiaojun Chen, Xingwei Wang, and Dacheng Tao.
\newblock Representation surgery for multi-task model merging.
\newblock {\em ICML}, 2024.

\bibitem{yangadamerging}
Enneng Yang, Zhenyi Wang, Li~Shen, Shiwei Liu, Guibing Guo, Xingwei Wang, and Dacheng Tao.
\newblock Adamerging: Adaptive model merging for multi-task learning.
\newblock In {\em The Twelfth International Conference on Learning Representations}, 2024.

\bibitem{yu2024unleashing}
Jun Yu, Yutong Dai, Xiaokang Liu, Jin Huang, Yishan Shen, Ke~Zhang, Rong Zhou, Eashan Adhikarla, Wenxuan Ye, Yixin Liu, et~al.
\newblock Unleashing the power of multi-task learning: A comprehensive survey spanning traditional, deep, and pretrained foundation model eras.
\newblock {\em arXiv preprint arXiv:2404.18961}, 2024.

\bibitem{yu2024language}
Le~Yu, Bowen Yu, Haiyang Yu, Fei Huang, and Yongbin Li.
\newblock Language models are super mario: Absorbing abilities from homologous models as a free lunch.
\newblock In {\em Forty-first International Conference on Machine Learning}, 2024.

\bibitem{zhou2024cross}
Zhanpeng Zhou, Zijun Chen, Yilan Chen, Bo~Zhang, and Junchi Yan.
\newblock Cross-task linearity emerges in the pretraining-finetuning paradigm.
\newblock {\em arXiv preprint arXiv:2402.03660}, 2024.

\end{thebibliography}
}

\end{document}